\documentclass[conference]{IEEEtran}

\usepackage{cite}

\usepackage[cmex10]{amsmath}
\usepackage{amsfonts}
\usepackage{amsthm}

\usepackage[table]{xcolor}
\usepackage{colortbl}
\usepackage{graphicx}

\allowdisplaybreaks

\IfFileExists{./.short}{
    \newcommand{\ifshortpaper}[1]{##1}
    \newcommand{\iffullpaper}[1]{}
}{
    \newcommand{\ifshortpaper}[1]{}
    \newcommand{\iffullpaper}[1]{##1}
}

\begin{document}

\title{An Asynchronous Implementation of\\the Limited Memory CMA-ES}
\author{
\IEEEauthorblockN{Viktor Arkhipov, Maxim Buzdalov, Anatoly Shalyto}
\IEEEauthorblockA{
ITMO University\\
49 Kronverkskiy prosp.\\
Saint-Petersburg, Russia, 197101\\
Email: \{arkhipov, buzdalov\}@rain.ifmo.ru, shalyto@mail.ifmo.ru
}
}

\maketitle

\setlength{\textfloatsep}{10pt plus 1.0pt minus 2.0pt}

\begin{abstract}
We present our asynchronous implementation of the LM-CMA-ES algorithm,
which is a modern evolution strategy for solving complex large-scale continuous optimization problems. 
Our implementation brings the best results when the number of cores is relatively high and the 
computational complexity of the fitness function is also high. 
The experiments with benchmark functions show that
it is able to overcome its origin 
on the Sphere function, reaches certain 
thresholds faster on the Rosenbrock and Ellipsoid function, and surprisingly performs much better 
than the original version on the Rastrigin function.
\end{abstract}

\section{Introduction}

Evolutionary algorithms are a nice tool for solving complex optimization problems for which 
no efficient algorithms are currently known.
They belong to black-box optimization algorithms, i.e.~the algorithms which learn 
the problem instance they solve by querying the function value, or ``fitness'', 
of a solution from that instance. In their design, evolutionary algorithms rely on the ideas 
which are rooted in natural evolution: problem solutions ``mutate'' (undergo small 
changes), ``mate'' or ``cross over'' (new solutions are created from parts of other solutions), 
undergo ``selection'' (only a part of solutions proceeds to the next iteration).

One of the most prominent features of evolutionary algorithms is the high degree of parallelism. 
Most evolutionary algorithms, such as genetic algorithms, evaluate hundreds of solutions in 
parallel, which allows efficient usage of multicore and distributed computer systems. 
However, many such algorithms are generational, that is, they must have many solutions 
to be completely evaluated before taking further actions. If the time of a single solution 
evaluation may differ from time to time, such evaluation scheme may be quite inefficient, 
because many threads finish their work too early and wait for long periods of time 
until the next work pieces become available. To overcome this problem, steady-state evolutionary
algorithms are developed, which do not have the notion of generations, so they have a potential 
to become asynchronous and to use computer resources in a better way.

Optimization problems are often subdivided into three groups: discrete problems, 
continuous problems and mixed problems. Among them, continuous problems received significant attention 
in classical mathematical analysis, so a rich body of methods was developed 
which makes optimization efficient by using information about gradient and second derivatives. 
Black-box algorithms for solving continuous optimization problems, while remaining in the 
black-box setting (i.e.~using only the function value, but not gradient, which may not exist at 
all), borrowed many ideas from classic methods. One of the most famous algorithms of this sort
is the Broyden-Fletcher-Goldfarb-Shanno algorithm, or simply BFGS~\cite{bfgs}. 
Modern evolution strategies, e.g. evolutionary algorithms for continuous optimization
which tend to self-adapt to the problem's properties, also follow this direction, 
which led to the appearance of the evolution strategy with covariance matrix adaptation 
(CMA-ES~\cite{hansen-2003}).

The CMA-ES algorithm, as follows from its name, learns a covariance matrix which encodes 
dependencies between decision variables, together with some more traditional parameters
like the step size. This algorithm shows very good convergence at many problems, 
including non-separable, multimodal and noisy problems. However, it comes with a drawback, 
which is high computational complexity ($O(n^2)$ per function evaluation, where $n$ is the problem size,
and $O(n^2)$ memory for storing the covariance matrix). An attempt to reduce memory requirements
and decrease the running time using numeric methods for Cholesky factorization was recently done in~\cite{numeric-cma-update}.
Several modifications were developed, including sep-CMA-ES~\cite{cma-es-diagonal}, 
which gives linear time and space complexity, but at the cost of worse performance on non-separable 
functions. However, most if not all algorithms from this family, while being generational by their 
nature, have a cost of maintaining the algorithm state after evaluation of the generation
which is comparable to or even higher than a single function evaluation. This makes it nearly 
impractical to use them in multicore or distributed computing environments,
especially when function evaluation is cheap.

The only attempt to turn an evolution strategy from the CMA-ES family into an asynchronous algorithm, 
which is known to authors, belongs to Tobias Glasmachers~\cite{glasmachers-async-nes}. 
This work, however, addresses a so-called \emph{natural evolution strategy}~\cite{natural-es},
which is much easier to make asynchronous due to properties of updates used in this algorithm. 
This approach can not be easily extended on other flavours of CMA-ES.

This paper presents our current research results on making asynchronous a more typical CMA 
algorithm, proposed by Ilya Loshchilov~\cite{lm-cma-es} and called ``limited memory CMA-ES'' (LM-CMA-ES).
This algorithm is designed to work with large scale optimization problems (with the number 
of dimensions $n \ge 1000$) which are not solvable with the usual CMA-ES. The key idea which made 
this possible is to compute the approximation of the product of a square root of the covariance 
matrix by a random vector sampled from the Gaussian distribution, which is a way CMA-ES samples new 
individuals, by an iterative process which references only $m$ vectors, $m \ll n$, from the past 
evaluations. The step sizes are restored in a similar way. This makes it possible to avoid storing 
the covariance matrix or its parts explicitly (which takes $O(n^2)$ memory) and finding its 
eigenpairs (which takes $O(n^3)$ time per generation). In the same time, this method does not 
sacrifice rotation symmetry in the way the sep-CMA-ES does.

The main idea of this paper is to change the algorithm which updates the stored data in order to 
make it incremental, which turns the LM-CMA-ES algorithm into a steady-state one.
After this change, the solutions can be evaluated asynchronously, which makes the overall 
implementation asynchronous. The results of the very early experimentation stage are presented in 
our previous paper~\cite{mendel-2015-async-cma}. Compared to that paper, we use more evaluation functions,
more algorithm configurations, measure much more data (not only CPU load)
and perform a comparison to the original LM-CMA-ES algorithm.

\iffullpaper{This is a full version of the paper which was accepted as a poster to the IEEE ICMLA conference in 2015.}

The rest of the paper is structured as follows. 
Section~\ref{algo-desc} contains the description of our asynchronous modification of the LM-CMA-ES.
Section~\ref{research-questions} sets up the research questions which are solved in the paper.
Section~\ref{experiments} describes the methodology used in experiment design, the experiment setup, 
results and their discussion.
Section~\ref{conclusion} concludes.

\section{Algorithm Description}\label{algo-desc}

In this section we describe the proposed asynchronous modification of the LM-CMA-ES algorithm.
Section~\ref{algo-desc-orig} briefly describes the original LM-CMA-ES algorithm as in~\cite{lm-cma-es},
and Section~\ref{algo-desc-diff} explains the modifications which we made.

\subsection{The Original LM-CMA-ES}\label{algo-desc-orig}

The key idea of the LM-CMA-ES algorithm is to implicitly restore the covariance matrix $C$ 
from $m \ll n$ selected vector pairs $(v^{(t)}, p_c^{(t)})$ sampled at some past moments of time. 
Each vector pair consists of $v^{(t)}$, the solution represented in the coordinate system 
based on eigenvectors of the covariance matrix $C^{(t)}$ corresponding to the moment of time $t$, 
and the evolution path $p_c^{(t)}$ represented in the same coordinate system.

The algorithm itself does not need the entire matrix by 
itself, but only its left Cholesky factor $A$ (such that $C = A \times A^T$) and its inverse $A^{-1}$. 
Generally they also require $\Theta(n^2)$ space to be stored, 
but if we know that the matrix $C$ is constructed from only $m$ pairs of vectors, we can use
these vectors to simulate an effect of $A \times x$ and $A^{-1} \times x$ for an arbitrary vector $x$. 
The total time needed to do this is $O(nm)$ per single function evaluation.

This algorithm, as any algorithm belonging to the CMA-ES family, has a number of parameters, 
such as $\lambda$, the number of sampled solutions per iteration, $\mu$, the number of best solutions 
which are used to construct the next reference search point, weights $w_i$ for creation of 
this search point, and a number of other parameters. However, they all have default values, 
which were shown to be efficient for optimization of multiple benchmark functions~\cite{lm-cma-es}.

\subsection{Our Modifications}\label{algo-desc-diff}

Our implementation is different from the LM-CMA-ES algorithm in the following key aspects:
\begin{itemize}
    \item We do not use generations of size $\lambda$. Instead, we use several threads which work 
    independently most of the time. First, a thread samples a solution based on the reference 
    search point, the implicit Cholesky factor $A$ and using generator of random numbers 
    from the normal distribution $\mathcal{N}(0,1)$. Then it evaluates this 
    solution using the fitness function. After that, the thread enters a critical section, performs
    update operations, makes copies of the necessary data and leaves the critical section.
    \item Due to the one-at-a-time semantics of an update, the modification no longer belongs to the
    $(\mu/\mu_w, \lambda)$ scheme, but rather can be described as $(m+1)$~-- the algorithm maintains 
    $m$~best solution vectors and updates this collection when a new solution comes. This may 
    have some drawbacks on complex functions, as the algorithm has a larger probability 
    to converge to a local optimum, but the simpler update scheme introduces smaller 
    computational costs inside a critical section, which improves running times.
\end{itemize}

\section{Research Questions}\label{research-questions}

How can we compare the quality of different algorithms, given a single problem to solve?
There are many measures which assist researchers in doing this. Probably the most three popular 
measures, especially when one considers multiple cores in use, are the following ones:
\begin{enumerate}
    \item The best possible solution quality. The main aim of this measure is to determine how 
          good the considered algorithm is in determining the problem's features, and which 
          fitness function values it is able to reach. To track this value, one can run the 
          algorithm until it finds an optimum within the given precision (for example, 
          $10^{-10}$), or until a sufficiently big computational budget is spent (for example, 
          $10^6$ fitness function evaluations), or until one can show that the algorithm cannot 
          find a better solution in a reasonable time.
    \item The best solution quality under a certain wall-clock time limit.
          This is quite a ``practical'' measure widely used in solving real-world problems.
          However it suffers a drawback: one cannot really compare the results of different 
          algorithms without any problem knowledge. For example, a difference of $0.1$ in the 
          fitness values can be really big for some problems and negligibly small for some other 
          problems.
    \item The wall-clock time to obtain a solution of the given quality. This is a more 
          ``theoretical'' measure because the computational budget is not limited, and it is often 
          a known optimum value which is taken to be a threshold. However we may expect a better 
          scaling of results achieved in this way, which depends less on the problem's properties.
\end{enumerate}

In this paper, we compare our asynchronous modification of the LM-CMA-ES algorithm with the 
original generational implementation of the same algorithm on different benchmark problems.
As these problems have varying difficulty for these algorithms, it is unrealistic to create the 
uniform experiment setups for all these problems. So we drop the second measure and consider only 
the first one (the best possible solution quality) and the third one (the wall-clock time to 
obtain a solution of the given quality). Based on these comparisons, we try to find answers for 
the following questions:
\begin{enumerate}
    \item Are the considered algorithms able to solve the problems to optimality (i.e. within the 
          precision of $10^{-10}$)? Which problems are tractable by which algorithms?
    \item What are the convergence properties of the algorithms?
          How the changes introduced in the asynchronous version affect the convergence?
    \item How the performance of the algorithms depends on the number of cores?
    \item How the performance of the algorithms depends on the computational complexity of the 
          fitness function?
    \item Which are the problems where the asynchronous version is better, and where is it worse?
\end{enumerate}

\section{Experiments}\label{experiments}

This section explains the methodology used while designing the 
experiments~(Section~\ref{experiments-methodology}),
shows the setup of the experiments, including technical details~(Section~\ref{experiments-setup}),
presents and explains the results of the preliminary experiments~(Section~\ref{experiments-prelim}),
and ends with the main experiment results and their discussion~(Section~\ref{experiments-results}).

\subsection{Methodology}\label{experiments-methodology}

When one configures an experiment, one has to choose one of the possible values for several parameters.
We consider the following important parameters:
\begin{enumerate}
    \item The problem type (the type of the function to be optimized),
    \item The problem dimension (the number of decision variables).
    \item The computational complexity of the problem 
          (the computational effort required to evaluate a single 
          solution, probably as a function of the problem dimension and the fitness value).
    \item The number of CPU cores used to run the algorithm.
\end{enumerate}

All these parameters, except for the third one, have been considered in the literature before. 
The third parameter (the computational complexity) typically was considered as a function of the 
problem dimension and the fitness value which is determined exclusively by the problem type. 
Due to the fourth research question, we decided to treat separately the computational complexity
because this gives a chance to adjust the ratio of fitness time to update time, 
which influences the overall CPU utilization, without influencing other problem characteristics. 
One can expect that, for example, as the fitness function becomes more and more ``heavy'', while 
all other parameters are left unchanged, the CPU utilization becomes better.

\subsection{Setup}\label{experiments-setup}

We used the following benchmark functions for the first experiment parameter:
\begin{itemize}
    \item the Sphere function:
          $f(x) = \sum_{i=1}^{n}{x_i^2}$;
    \item the Rastrigin function:\\
          $f(x) = 10n + \sum_{i=1}^{n}{\left(x_i^2 - 10 \cos (2\pi x_i)\right)}$;
    \item the Rosenbrock function:\\
          $f(x) = \sum_{i=1}^{n-1}\left(100 \left(x_{i+1} - x_i^2\right)^2 + (1 - x_i)^2\right)$;
    \item the Ellipsoid function:
          $f(x) = \sum_{i=1}^{n}{\left(\sum_{j=1}^{i}{x_j}\right)^2}$.
\end{itemize}

For the second experiment parameter (the problem's dimension), we use the values 100, 300 and 1000.

For the third experiment parameter (the computational complexity of a fitness function), we either 
use the fitness function as it is, or augment it with a procedure which changes nothing in the 
fitness function result but introduces additional computation complexity. 
This additional complexity is 
achieved by running the Bellman-Ford algorithm on a randomly generated graph with 10\,000 edges and
200, 400, or 600 vertices. This produces the linear growth of the additional complexity, which is 
similar to or exceeds the ``natural'' computational complexity of the fitness function.

For the fourth experiment parameter (the number of cores), we use the values 2, 4, 6, and 8.

There were three termination criteria, which stop the algorithm under the following conditions:
\begin{enumerate}
    \item The function value is close to the optimum with the absolute precision of 
          $10^{-10}$. The optimum values of all the considered functions are known to be zeros.
    \item The number of fitness function evaluations exceeds $10^6$.
    \item For the asynchronous implementation, the value of the parameter $\sigma$ becomes less 
          than $10^{-20}$, which reflects the algorithm's stagnation in the current point.
\end{enumerate}

Each experiment configuration was run for 100 times for configurations without 
additional computational complexity and for 50 times otherwise.

For monitoring the overall performance, both in terms of convergence and of using multiple 
cores, we track the median values of the wall-clock running time needed to reach a certain 
threshold $t$, and the median values of the processor's cores load for the time from the beginning 
of the run until the threshold $t$. For each problem we have to choose several values of the 
threshold $t$, which we do by analysing the preliminary experiments.

\subsection{Preliminary Experiments and Their Results}\label{experiments-prelim}

The preliminary experiments were designed to get preliminary answers to the first two research 
questions, and also to find suitable fitness function thresholds for each fitness function we 
used. We did several trial runs of the asynchronous implementation of the LM-CMA-ES algorithm
with the problem dimension of 100 without additional computation complexity and with different 
numbers of available cores.

\begin{figure}[!t]
\includegraphics[width=\columnwidth]{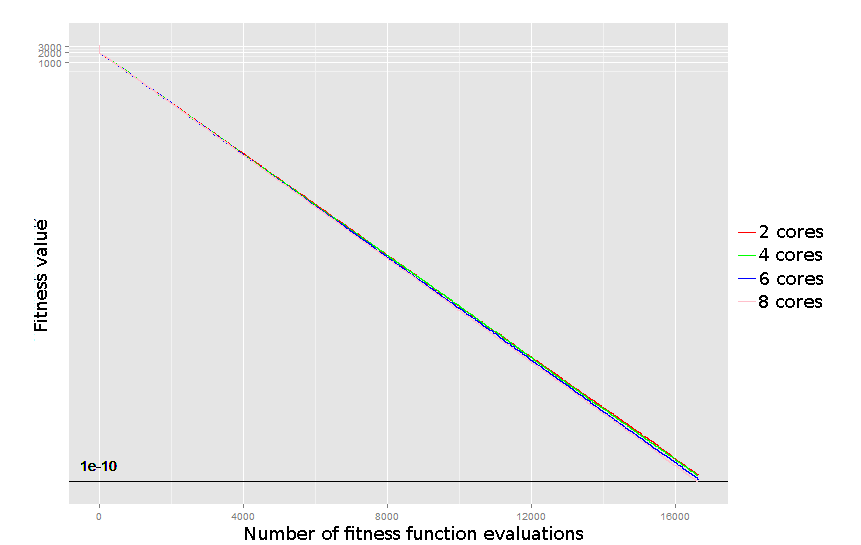}
\caption{Example fitness function plots on Sphere}\label{preliminary-sphere}
\end{figure}

\begin{figure}[!t]
\includegraphics[width=\columnwidth]{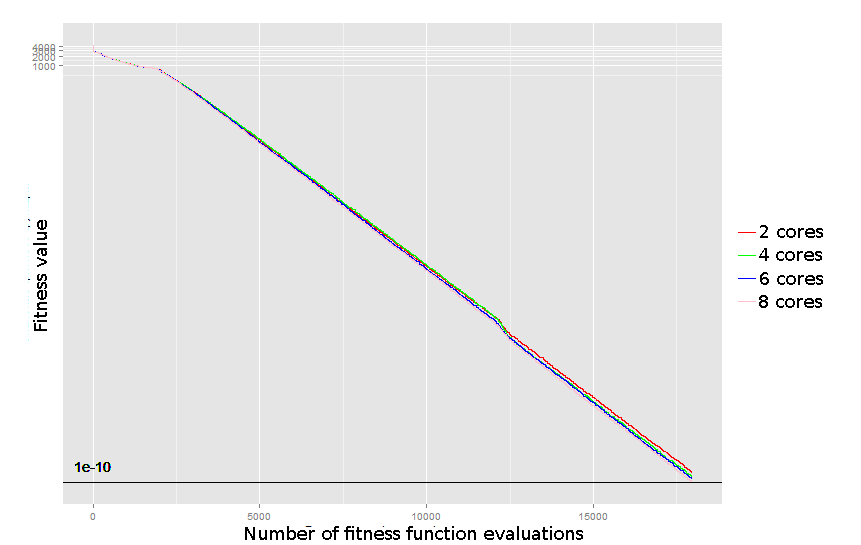}
\caption{Example fitness function plots on Rastrigin}\label{preliminary-rastrigin}
\end{figure}

Fig.~\ref{preliminary-sphere} and~\ref{preliminary-rastrigin} show that the Sphere and Rastrigin 
functions are typically optimized to the global optimum independently of the number of cores
(in other words, the algorithm always finds a point with the fitness value less than $10^{-10}$).
One can notice the linear convergence in the case of these two functions. For the Rastrigin function
there is one linear piece with a smaller slope in the beginning, which seems to correspond to the 
search for the surroundings of the global optimum. For these two functions the 
logarithmically equidistant thresholds seem to be the optimal choice, so we select the values of
$1$, $3 \cdot 10^{-2}$, $10^{-5}$, $3 \cdot 10^{-7}$, $10^{-10}$.

On Fig.~\ref{preliminary-rosenbrock} one can see a different kind of convergence for the 
Rosenbrock function. This function is quite difficult for the asynchronous algorithm,
as all the runs converge to approximately 90 and then stop having any progress. This is a 
characteristic trait of this function, as the optimizers typically reach the valley first and then 
try to find an optimum within the valley using a very small gradient. For this function we take 
as a threshold the value of 100, which is close enough to the stagnation point, and the value of 
3000, which should be reachable by any sensible optimizer. We also include the value $10^{-10}$ 
from the termination criterion and two intermediate values of $10^{-2}$ and $10^{-6}$.

\begin{figure}[!t]
\includegraphics[width=\columnwidth]{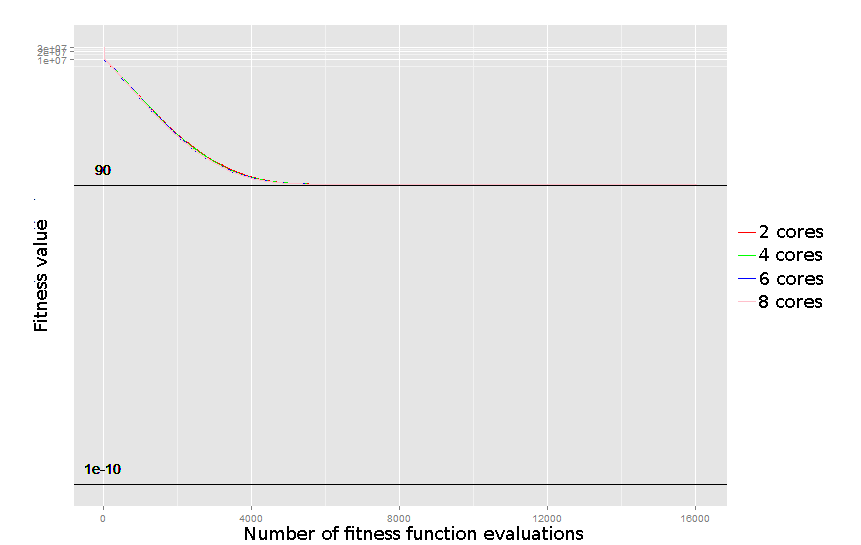}
\caption{Example fitness function plots on Rosenbrock}\label{preliminary-rosenbrock}
\end{figure}

\begin{figure}[!t]
\includegraphics[width=\columnwidth]{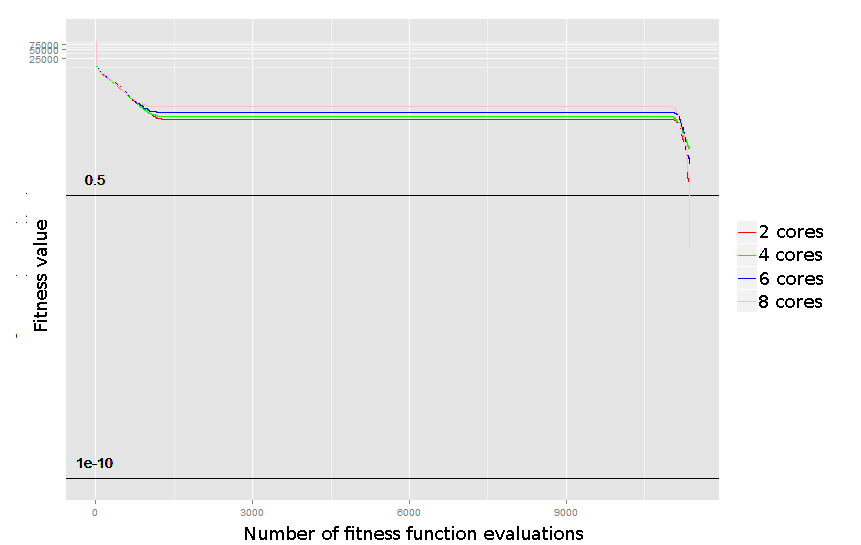}
\caption{Example fitness function plots on Ellipsoid}\label{preliminary-ellipsoid}
\end{figure}

Finally, Fig.~\ref{preliminary-ellipsoid} shows a similar but slightly different kind of 
convergence for the Ellipsoid function. In this case, unlike Rosenbrock, after a large number of 
fitness function evaluations the algorithm is sometimes able to learn a new gradient and find the 
optimum. This can follow from the 1/5-th rule used in the algorithm, which is not very efficient 
in this problem. For the threshold values, we took the same values as for the Rosenbrock function.

\subsection{Main Experiment Results and Discussion}\label{experiments-results}

\ifshortpaper{
    The results are presented in Tables~\ref{main-sphere}--\ref{main-ellipsoid}.
    For the sake of saving place, only the results for additional complexity of 0 and 600 are presented,
    and the dimension 300 is omitted. The full version of the paper is available at ArXiV~\footnote{??}.

    \input{tables-short.tex}
}
\iffullpaper{
    The results are presented in Tables~\ref{main-sphere}--\ref{main-ellipsoid}.
\begin{table*}[t]
\begin{center}
\caption{
Experiment results for the Sphere function.
Legend: ``Proposed'' is the asynchronous modification of LM-CMA-ES described in this paper,
``LM-CMA'' is the original LM-CMA-ES implementation. The table entries consist of
the median wall-clock time to reach the corresponding threshold (in seconds)
and of the median processor load value, separated by a forward slash.
}\label{main-sphere}
\begin{tabular}{cc|ccccc|ccccc}\hline
n & Algo & \multicolumn{5}{c|}{Thresholds} & \multicolumn{5}{c}{Thresholds} \\\cline{3-12}
 & & $1$ & $3 \cdot 10^{-2}$ & $10^{-5}$ & $3 \cdot 10^{-7}$ & $10^{-10}$ & $1$ & $3 \cdot 10^{-2}$ & $10^{-5}$ & $3 \cdot 10^{-7}$ & $10^{-10}$ \\\hline
  &      & \multicolumn{5}{c|}{2 cores, additional complexity 0} & \multicolumn{5}{c}{2 cores, additional complexity 200} \\\hline
100 & Proposed & 0.17/1.6 & 0.25/1.6 & 0.42/1.6 & 0.49/1.6 & 0.65/1.6 & 0.46/1.9 & 0.62/1.9 & 1.07/1.9 & 1.28/1.9 & 1.69/1.9 \\
 & LM-CMA & 0.11/1.7 & 0.15/1.7 & 0.24/1.7 & 0.28/1.7 & 0.38/1.7 & 0.26/1.9 & 0.36/1.9 & 0.6/1.9 & 0.71/1.9 & 0.96/1.9 \\
\hline
300 & Proposed & 1.18/1.7 & 1.63/1.7 & 2.67/1.7 & 3.13/1.7 & 4.15/1.7 & 1.72/1.8 & 2.39/1.8 & 3.86/1.8 & 4.52/1.8 & 6.03/1.8 \\
 & LM-CMA & 0.55/1.7 & 0.76/1.8 & 1.22/1.8 & 1.42/1.8 & 1.91/1.8 & 0.97/1.8 & 1.32/1.8 & 2.15/1.8 & 2.5/1.8 & 3.33/1.8 \\
\hline
1000 & Proposed & 10.38/1.6 & 14.04/1.6 & 22.22/1.6 & 25.75/1.6 & 33.84/1.6 & 10.46/1.7 & 13.98/1.7 & 22.11/1.7 & 25.67/1.7 & 33.75/1.7 \\
 & LM-CMA & 5.45/1.8 & 7.42/1.8 & 11.87/1.8 & 13.88/1.8 & 18.34/1.8 & 6.2/1.8 & 8.56/1.8 & 13.88/1.8 & 16.43/1.8 & 22.08/1.8 \\
\hline
  &      & \multicolumn{5}{c|}{2 cores, additional complexity 400} & \multicolumn{5}{c}{2 cores, additional complexity 600} \\\hline
100 & Proposed & 1.59/1.9 & 2.23/1.9 & 3.74/1.9 & 4.4/1.9 & 5.94/1.9 & 3.23/1.9 & 4.78/1.9 & 8.19/1.9 & 9.49/1.9 & 12.96/1.9 \\
 & LM-CMA & 1.45/1.9 & 2.56/1.9 & 2.97/1.9 & 4.04/1.9 & 1.01/1.9 & 2.29/1.9 & 3.39/1.9 & 5.66/1.9 & 6.8/1.9 & 9.39/1.9 \\
\hline
300 & Proposed & 3.87/1.9 & 5.4/1.9 & 8.79/1.9 & 10.31/1.9 & 13.82/1.9 & 9.27/1.9 & 12.99/1.9 & 21.32/1.9 & 25.49/1.9 & 34.05/1.9 \\
 & LM-CMA & 3.02/1.9 & 4.26/1.9 & 7.02/1.9 & 8.18/1.9 & 10.91/1.9 & 6.31/1.9 & 8.79/1.9 & 14.45/1.9 & 16.84/1.9 & 22.68/1.9 \\
\hline
1000 & Proposed & 17.3/1.8 & 23.46/1.8 & 36.71/1.8 & 42.81/1.8 & 56.0/1.8 & 30.73/1.9 & 41.52/1.9 & 65.3/1.9 & 76.12/1.9 & 100.51/1.9 \\
 & LM-CMA & 14.73/1.8 & 20.14/1.8 & 32.7/1.8 & 38.11/1.8 & 52.13/1.8 & 23.79/1.9 & 32.4/1.9 & 51.3/1.9 & 60.22/1.9 & 79.74/1.9 \\
\hline
  &      & \multicolumn{5}{c|}{4 cores, additional complexity 0} & \multicolumn{5}{c}{4 cores, additional complexity 200} \\\hline
100 & Proposed & 0.27/1.1 & 0.39/1.2 & 0.66/1.1 & 0.76/1.1 & 1.03/1.1 & 0.22/3.5 & 0.31/3.5 & 0.54/3.4 & 0.64/3.4 & 0.86/3.4 \\
 & LM-CMA & 0.07/2.5 & 0.1/2.5 & 0.17/2.5 & 0.2/2.5 & 0.27/2.5 & 0.16/3.4 & 0.23/3.4 & 0.39/3.5 & 0.46/3.5 & 0.66/3.5 \\
\hline
300 & Proposed & 1.42/1.7 & 1.96/1.7 & 3.2/1.7 & 3.74/1.7 & 4.99/1.6 & 1.09/3.0 & 1.53/3.0 & 2.53/3.0 & 2.98/3.0 & 3.97/3.0 \\
 & LM-CMA & 0.43/2.6 & 0.59/2.6 & 0.95/2.6 & 1.11/2.6 & 1.48/2.6 & 0.65/3.4 & 0.93/3.4 & 1.49/3.4 & 1.75/3.4 & 2.33/3.4 \\
\hline
1000 & Proposed & 10.31/1.9 & 13.83/1.9 & 22.03/1.9 & 25.52/1.9 & 33.62/1.9 & 7.93/2.5 & 10.69/2.5 & 16.98/2.5 & 19.75/2.5 & 25.95/2.5 \\
 & LM-CMA & 3.77/2.8 & 5.13/2.8 & 8.2/2.8 & 9.52/2.8 & 12.57/2.8 & 4.4/3.0 & 5.96/3.0 & 9.58/3.0 & 11.28/3.0 & 14.83/3.0 \\
\hline
  &      & \multicolumn{5}{c|}{4 cores, additional complexity 400} & \multicolumn{5}{c}{4 cores, additional complexity 600} \\\hline
100 & Proposed & 0.66/3.9 & 0.94/3.9 & 1.59/3.9 & 1.88/3.9 & 2.5/3.9 & 1.31/3.9 & 1.92/3.9 & 3.27/3.9 & 3.81/3.9 & 5.2/3.9 \\
 & LM-CMA & 0.73/3.6 & 1.06/3.6 & 1.76/3.6 & 2.13/3.6 & 2.84/3. & 1.51/3.7 & 2.14/3.7 & 3.64/3.7 & 4.36/3.7 & 5.67/3.7 \\
\hline
300 & Proposed & 2.04/3.7 & 2.85/3.7 & 4.7/3.7 & 5.5/3.7 & 7.27/3.7 & 4.32/3.9 & 6.03/3.9 & 10.03/3.9 & 11.89/3.9 & 15.92/3.9 \\
 & LM-CMA & 2.08/3.6 & 2.88/3.6 & 4.9/3.6 & 5.69/3.6 & 7.49/3.6 & 4.58/3.7 & 6.56/3.7 & 10.88/3.7 & 12.89/3.7 & 17.51/3.7 \\
\hline
1000 & Proposed & 9.9/3.0 & 13.41/3.0 & 21.28/3.0 & 24.84/3.0 & 32.64/3.0 & 16.55/3.6 & 22.38/3.6 & 34.93/3.6 & 40.52/3.6 & 53.01/3.6 \\
 & LM-CMA & 8.82/3.2 & 12.03/3.2 & 19.97/3.2 & 23.54/3.2 & 31.36/3.2 & 15.15/3.5 & 20.28/3.5 & 33.64/3.5 & 39.07/3.5 & 52.54/3.5 \\
\hline
  &      & \multicolumn{5}{c|}{6 cores, additional complexity 0} & \multicolumn{5}{c}{6 cores, additional complexity 200} \\\hline
100 & Proposed & 0.29/1.0 & 0.42/1.0 & 0.7/0.9 & 0.84/0.9 & 1.12/0.9 & 0.28/4.4 & 0.37/4.4 & 0.61/4.4 & 0.72/4.4 & 0.93/4.3 \\
 & LM-CMA & 0.08/2.3 & 0.11/2.4 & 0.17/2.3 & 0.21/2.3 & 0.28/2.3 & 0.16/4.7 & 0.23/4.8 & 0.37/4.8 & 0.43/4.8 & 0.57/4.8 \\
\hline
300 & Proposed & 1.6/1.3 & 2.24/1.3 & 3.65/1.3 & 4.29/1.3 & 5.7/1.3 & 1.22/3.3 & 1.69/3.3 & 2.77/3.4 & 3.2/3.4 & 4.27/3.4 \\
 & LM-CMA & 0.44/2.5 & 0.6/2.5 & 0.99/2.5 & 1.15/2.5 & 1.54/2.5 & 0.51/4.6 & 0.69/4.7 & 1.16/4.8 & 1.36/4.7 & 1.82/4.8 \\
\hline
1000 & Proposed & 11.08/1.7 & 14.95/1.7 & 23.6/1.7 & 27.49/1.7 & 36.26/1.7 & 8.83/2.5 & 11.9/2.5 & 18.87/2.5 & 21.96/2.5 & 29.04/2.5 \\
 & LM-CMA & 3.55/3.0 & 4.86/3.1 & 7.76/3.1 & 9.02/3.1 & 11.89/3.1 & 4.34/3.4 & 5.96/3.5 & 9.45/3.5 & 10.96/3.5 & 14.52/3.4 \\
\hline
  &      & \multicolumn{5}{c|}{6 cores, additional complexity 400} & \multicolumn{5}{c}{6 cores, additional complexity 600} \\\hline
100 & Proposed & 0.48/5.7 & 0.7/5.8 & 1.14/5.8 & 1.36/5.8 & 1.88/5.8 & 0.99/5.9 & 1.46/5.9 & 2.47/5.9 & 2.94/5.9 & 4.02/5.9 \\
 & LM-CMA & 0.63/5.2 & 0.86/5.2 & 1.43/5.2 & 1.67/5.2 & 2.2/5.1 & 1.42/5.3 & 2.05/5.3 & 3.2/5.3 & 3.73/5.3 & 4.94/5.2 \\
\hline
300 & Proposed & 1.57/5.1 & 2.23/5.1 & 3.67/5.1 & 4.33/5.1 & 5.81/5.1 & 3.3/5.7 & 4.52/5.7 & 7.49/5.7 & 8.82/5.7 & 11.87/5.7 \\
 & LM-CMA & 1.49/5.1 & 2.12/5.1 & 3.55/5.1 & 4.13/5.1 & 5.52/5.1 & 3.52/5.3 & 4.83/5.3 & 7.85/5.2 & 9.28/5.2 & 12.45/5.2 \\
\hline
1000 & Proposed & 9.38/4.0 & 12.56/4.0 & 19.8/4.0 & 23.04/4.0 & 30.46/4.0 & 12.66/4.9 & 16.99/4.9 & 26.84/4.9 & 31.11/4.9 & 40.95/4.9 \\
 & LM-CMA & 8.03/4.0 & 11.44/4.0 & 18.44/3.9 & 21.55/3.9 & 29.44/3.9 & 13.02/4.8 & 17.75/4.8 & 29.68/4.8 & 34.6/4.8 & 46.78/4.8 \\
\hline
  &      & \multicolumn{5}{c|}{8 cores, additional complexity 0} & \multicolumn{5}{c}{8 cores, additional complexity 200} \\\hline
100 & Proposed & 0.3/1.0 & 0.43/1.0 & 0.73/1.0 & 0.86/1.0 & 1.16/1.0 & 0.32/4.1 & 0.46/4.1 & 0.76/4.1 & 0.9/4.2 & 1.18/4.2 \\
 & LM-CMA & 0.08/2.3 & 0.11/2.3 & 0.18/2.3 & 0.21/2.3 & 0.28/2.3 & 0.13/5.3 & 0.2/5.3 & 0.35/5.2 & 0.4/5.3 & 0.51/5.3 \\
\hline
300 & Proposed & 1.64/1.4 & 2.27/1.4 & 3.73/1.4 & 4.37/1.3 & 5.83/1.3 & 1.39/3.2 & 1.97/3.2 & 3.16/3.3 & 3.71/3.2 & 4.95/3.2 \\
 & LM-CMA & 0.46/2.4 & 0.64/2.4 & 1.03/2.4 & 1.2/2.4 & 1.6/2.4 & 0.42/5.6 & 0.58/5.7 & 0.97/5.7 & 1.13/5.7 & 1.49/5.7 \\
\hline
1000 & Proposed & 11.18/1.7 & 14.99/1.7 & 23.83/1.7 & 27.69/1.7 & 36.45/1.7 & 9.22/2.5 & 12.56/2.5 & 19.9/2.5 & 23.13/2.5 & 30.47/2.5 \\
 & LM-CMA & 3.63/3.0 & 4.94/3.0 & 7.91/3.0 & 9.22/3.0 & 12.15/3.0 & 4.72/3.6 & 6.47/3.7 & 10.77/3.6 & 12.28/3.6 & 16.05/3.6 \\
\hline
  &      & \multicolumn{5}{c|}{8 cores, additional complexity 400} & \multicolumn{5}{c}{8 cores, additional complexity 600} \\\hline
100 & Proposed & 0.39/7.3 & 0.55/7.4 & 0.93/7.4 & 1.09/7.4 & 1.46/7.3 & 1.19/7.8 & 1.96/7.8 & 2.3/7.8 & 3.12/7.8 & 0.85/7.8 \\
 & LM-CMA & 0.49/6.6 & 0.68/6.5 & 1.13/6.5 & 1.32/6.5 & 1.77/6.5 & 1.28/6.8 & 1.73/6.8 & 2.96/6.8 & 3.47/6.8 & 4.52/6.8 \\
\hline
300 & Proposed & 1.54/6.6 & 2.12/6.6 & 3.53/6.5 & 4.08/6.5 & 5.39/6.6 & 2.69/7.4 & 3.74/7.5 & 6.25/7.5 & 7.46/7.5 & 10.07/7.5 \\
 & LM-CMA & 1.23/6.5 & 1.79/6.6 & 3.02/6.6 & 3.51/6.6 & 4.79/6.6 & 2.87/6.8 & 3.87/6.8 & 6.34/6.8 & 7.35/6.8 & 9.93/6.8 \\
\hline
1000 & Proposed & 9.41/4.4 & 12.65/4.4 & 20.17/4.4 & 23.59/4.4 & 31.05/4.4 & 11.86/6.1 & 16.07/6.1 & 25.53/6.1 & 29.48/6.1 & 39.06/6.1 \\
 & LM-CMA & 8.47/4.4 & 11.6/4.4 & 18.91/4.4 & 22.32/4.3 & 29.63/4.2 & 12.41/5.9 & 17.39/5.8 & 28.12/5.8 & 32.75/5.8 & 43.58/5.8 \\
\hline
\end{tabular}
\end{center}
\end{table*}
\begin{table*}[t]
\begin{center}
\caption{
Experiment results for the Rastrigin function.
Legend: ``Proposed'' is the asynchronous modification of LM-CMA-ES described in this paper,
``LM-CMA'' is the original LM-CMA-ES implementation. The table entries consist of
the median wall-clock time to reach the corresponding threshold (in seconds)
and of the median processor load value, separated by a forward slash.
}\label{main-rastrigin}
\begin{tabular}{cc|ccccc|ccccc}\hline
n & Algo & \multicolumn{5}{c|}{Thresholds} & \multicolumn{5}{c}{Thresholds} \\\cline{3-12}
 & & $1$ & $3 \cdot 10^{-2}$ & $10^{-5}$ & $3 \cdot 10^{-7}$ & $10^{-10}$ & $1$ & $3 \cdot 10^{-2}$ & $10^{-5}$ & $3 \cdot 10^{-7}$ & $10^{-10}$ \\\hline
  &      & \multicolumn{5}{c|}{2 cores, additional complexity 0} & \multicolumn{5}{c}{2 cores, additional complexity 200} \\\hline
100 & Proposed & 0.29/1.6 & 0.37/1.6 & 0.56/1.6 & 0.64/1.6 & 0.83/1.6 & 0.68/1.9 & 0.91/1.9 & 1.41/1.9 & 1.63/1.9 & 2.16/1.9 \\
 & LM-CMA & --- & --- & --- & --- & --- & --- & --- & --- & --- & --- \\
\hline
300 & Proposed & 1.97/1.7 & 2.51/1.7 & 3.72/1.7 & 4.24/1.7 & 5.41/1.7 & 2.53/1.8 & 3.18/1.8 & 4.74/1.8 & 5.45/1.8 & 6.93/1.8 \\
 & LM-CMA & --- & --- & --- & --- & --- & --- & --- & --- & --- & --- \\
\hline
1000 & Proposed & 17.54/1.6 & 21.67/1.6 & 31.17/1.6 & 35.24/1.6 & 44.13/1.6 & 18.83/1.7 & 23.31/1.7 & 33.68/1.7 & 38.37/1.7 & 48.24/1.7 \\
 & LM-CMA & --- & --- & --- & --- & --- & --- & --- & --- & --- & --- \\
\hline
  &      & \multicolumn{5}{c|}{2 cores, additional complexity 400} & \multicolumn{5}{c}{2 cores, additional complexity 600} \\\hline
100 & Proposed & 2.26/1.9 & 3.0/1.9 & 4.72/1.9 & 5.53/1.9 & 7.12/1.9 & 5.07/1.9 & 6.67/1.9 & 10.4/1.9 & 12.11/1.9 & 15.81/1.9 \\
 & LM-CMA & --- & --- & --- & --- & --- & --- & --- & --- & --- & --- \\
\hline
300 & Proposed & 6.76/1.9 & 8.6/1.9 & 12.74/1.9 & 14.62/1.9 & 18.89/1.9 & 12.32/1.9 & 15.85/1.9 & 23.79/1.9 & 27.34/1.9 & 35.07/1.9 \\
 & LM-CMA & --- & --- & --- & --- & --- & --- & --- & --- & --- & --- \\
\hline
1000 & Proposed & 32.42/1.8 & 40.52/1.8 & 58.98/1.8 & 67.34/1.8 & 85.16/1.8 & 52.96/1.9 & 66.23/1.9 & 95.9/1.9 & 109.09/1.9 & 136.96/1.9 \\
 & LM-CMA & --- & --- & --- & --- & --- & --- & --- & --- & --- & --- \\
\hline
  &      & \multicolumn{5}{c|}{4 cores, additional complexity 0} & \multicolumn{5}{c}{4 cores, additional complexity 200} \\\hline
100 & Proposed & 0.38/1.9 & 0.49/1.8 & 0.74/1.7 & 0.85/1.7 & 1.12/1.7 & 0.44/3.4 & 0.59/3.3 & 0.8/3.2 & 0.9/3.2 & 1.17/3.2 \\
 & LM-CMA & --- & --- & --- & --- & --- & --- & --- & --- & --- & --- \\
\hline
300 & Proposed & 1.84/2.3 & 2.35/2.3 & 3.5/2.2 & 4.0/2.2 & 5.14/2.2 & 1.54/3.0 & 2.0/3.0 & 3.06/3.0 & 3.49/3.0 & 4.43/3.0 \\
 & LM-CMA & --- & --- & --- & --- & --- & --- & --- & --- & --- & --- \\
\hline
1000 & Proposed & 14.14/2.3 & 17.59/2.3 & 25.5/2.3 & 28.93/2.3 & 36.26/2.3 & 11.6/2.6 & 14.46/2.6 & 21.18/2.6 & 24.16/2.6 & 30.27/2.6 \\
 & LM-CMA & --- & --- & --- & --- & --- & --- & --- & --- & --- & --- \\
\hline
  &      & \multicolumn{5}{c|}{4 cores, additional complexity 400} & \multicolumn{5}{c}{4 cores, additional complexity 600} \\\hline
100 & Proposed & 1.0/3.8 & 1.3/3.8 & 1.96/3.8 & 2.23/3.8 & 2.85/3.8 & 1.98/3.9 & 2.53/3.9 & 3.77/3.8 & 4.31/3.9 & 5.49/3.9 \\
 & LM-CMA & --- & --- & --- & --- & --- & --- & --- & --- & --- & --- \\
\hline
300 & Proposed & 2.88/3.7 & 3.7/3.6 & 5.46/3.6 & 6.31/3.6 & 8.04/3.7 & 5.79/3.8 & 7.37/3.8 & 11.14/3.8 & 12.79/3.8 & 16.53/3.8 \\
 & LM-CMA & --- & --- & --- & --- & --- & --- & --- & --- & --- & --- \\
\hline
1000 & Proposed & 14.71/3.1 & 18.36/3.1 & 27.13/3.0 & 30.76/3.0 & 38.65/3.0 & 22.88/3.5 & 28.49/3.5 & 40.98/3.5 & 46.64/3.5 & 58.23/3.5 \\
 & LM-CMA & --- & --- & --- & --- & --- & --- & --- & --- & --- & --- \\
\hline
  &      & \multicolumn{5}{c|}{6 cores, additional complexity 0} & \multicolumn{5}{c}{6 cores, additional complexity 200} \\\hline
100 & Proposed & 0.48/1.1 & 0.63/1.1 & 0.96/1.0 & 1.11/1.0 & 1.43/1.0 & 0.43/3.1 & 0.55/3.1 & 0.83/3.2 & 0.98/3.1 & 1.32/3.0 \\
 & LM-CMA & --- & --- & --- & --- & --- & --- & --- & --- & --- & --- \\
\hline
300 & Proposed & 2.47/1.6 & 3.17/1.6 & 4.72/1.5 & 5.4/1.5 & 6.93/1.5 & 1.59/3.8 & 2.04/3.7 & 3.1/3.6 & 3.54/3.6 & 4.54/3.7 \\
 & LM-CMA & --- & --- & --- & --- & --- & --- & --- & --- & --- & --- \\
\hline
1000 & Proposed & 15.9/2.1 & 19.9/2.1 & 28.98/2.0 & 32.92/2.0 & 41.43/2.0 & 13.12/2.7 & 16.49/2.6 & 24.01/2.6 & 27.29/2.6 & 34.32/2.6 \\
 & LM-CMA & --- & --- & --- & --- & --- & --- & --- & --- & --- & --- \\
\hline
  &      & \multicolumn{5}{c|}{6 cores, additional complexity 400} & \multicolumn{5}{c}{6 cores, additional complexity 600} \\\hline
100 & Proposed & 0.7/5.6 & 0.92/5.6 & 1.33/5.6 & 1.55/5.6 & 2.04/5.6 & 1.34/5.8 & 1.76/5.8 & 2.62/5.7 & 3.09/5.7 & 4.24/5.7 \\
 & LM-CMA & --- & --- & --- & --- & --- & --- & --- & --- & --- & --- \\
\hline
300 & Proposed & 2.22/5.0 & 2.9/5.1 & 4.3/5.1 & 4.93/5.1 & 6.35/5.2 & 4.16/5.6 & 5.38/5.6 & 8.12/5.6 & 9.36/5.6 & 12.1/5.6 \\
 & LM-CMA & --- & --- & --- & --- & --- & --- & --- & --- & --- & --- \\
\hline
1000 & Proposed & 13.72/3.8 & 17.22/3.8 & 25.13/3.8 & 28.71/3.8 & 36.15/3.8 & 17.74/4.8 & 22.19/4.7 & 32.23/4.7 & 36.81/4.7 & 46.23/4.7 \\
 & LM-CMA & --- & --- & --- & --- & --- & --- & --- & --- & --- & --- \\
\hline
  &      & \multicolumn{5}{c|}{8 cores, additional complexity 0} & \multicolumn{5}{c}{8 cores, additional complexity 200} \\\hline
100 & Proposed & 0.47/1.0 & 0.61/1.0 & 0.94/0.9 & 1.09/0.9 & 1.41/0.9 & 0.6/2.3 & 0.81/2.3 & 1.23/2.3 & 1.4/2.3 & 1.78/2.5 \\
 & LM-CMA & --- & --- & --- & --- & --- & --- & --- & --- & --- & --- \\
\hline
300 & Proposed & 2.58/1.4 & 3.26/1.4 & 4.85/1.4 & 5.57/1.4 & 7.21/1.3 & 2.15/2.9 & 2.74/2.8 & 4.14/2.7 & 4.74/2.7 & 6.17/2.7 \\
 & LM-CMA & --- & --- & --- & --- & --- & --- & --- & --- & --- & --- \\
\hline
1000 & Proposed & 16.35/2.0 & 20.47/1.9 & 29.7/1.9 & 33.83/1.9 & 42.39/1.9 & 14.83/2.3 & 18.53/2.3 & 27.27/2.3 & 30.8/2.3 & 38.83/2.3 \\
 & LM-CMA & --- & --- & --- & --- & --- & --- & --- & --- & --- & --- \\
\hline
  &      & \multicolumn{5}{c|}{8 cores, additional complexity 400} & \multicolumn{5}{c}{8 cores, additional complexity 600} \\\hline
100 & Proposed & 0.7/5.4 & 0.92/5.5 & 1.4/5.5 & 1.61/5.6 & 2.11/5.8 & 1.12/7.5 & 1.5/7.4 & 2.38/7.4 & 2.75/7.5 & 3.56/7.4 \\
 & LM-CMA & --- & --- & --- & --- & --- & --- & --- & --- & --- & --- \\
\hline
300 & Proposed & 2.25/5.3 & 2.87/5.3 & 4.35/5.3 & 4.97/5.3 & 6.4/5.2 & 4.01/7.1 & 5.07/7.1 & 7.37/7.0 & 8.41/7.1 & 10.88/7.1 \\
 & LM-CMA & --- & --- & --- & --- & --- & --- & --- & --- & --- & --- \\
\hline
1000 & Proposed & 14.23/4.2 & 17.73/4.1 & 25.76/4.1 & 29.27/4.1 & 36.93/4.1 & 18.1/5.7 & 22.67/5.7 & 33.06/5.7 & 37.62/5.7 & 46.82/5.6 \\
 & LM-CMA & --- & --- & --- & --- & --- & --- & --- & --- & --- & --- \\
\hline
\end{tabular}
\end{center}
\end{table*}
\begin{table*}[t]
\begin{center}
\caption{
Experiment results for the Rosenbrock function.
Legend: ``Proposed'' is the asynchronous modification of LM-CMA-ES described in this paper,
``LM-CMA'' is the original LM-CMA-ES implementation. The table entries consist of
the median wall-clock time to reach the corresponding threshold (in seconds)
and of the median processor load value, separated by a forward slash.
}\label{main-rosenbrock}
\begin{tabular}{cc|ccccc|ccccc}\hline
n & Algo & \multicolumn{5}{c|}{Thresholds} & \multicolumn{5}{c}{Thresholds} \\\cline{3-12}
 & & $3000$ & $100$ & $10^{-2}$ & $3 \cdot 10^{-6}$ & $10^{-10}$ & $3000$ & $100$ & $10^{-2}$ & $3 \cdot 10^{-6}$ & $10^{-10}$ \\\hline
  &      & \multicolumn{5}{c|}{2 cores, additional complexity 0} & \multicolumn{5}{c}{2 cores, additional complexity 200} \\\hline
100 & Proposed & 0.09/1.6 & 0.23/1.6 & --- & --- & --- & 0.29/1.9 & 0.71/1.9 & --- & --- & --- \\
 & LM-CMA & 0.05/1.7 & 1.32/1.7 & 6.41/1.7 & 6.55/1.7 & 6.72/1.7 & 0.16/1.8 & 1.9/1.9 & 20.5/1.8 & 21.07/1.8 & 21.61/1.8 \\
\hline
300 & Proposed & 0.71/1.6 & --- & --- & --- & --- & 1.08/1.8 & --- & --- & --- & --- \\
 & LM-CMA & 0.43/1.8 & 47.04/1.7 & 53.22/1.7 & 54.03/1.7 & 54.7/1.7 & 0.72/1.8 & 69.03/1.8 & 23.27/0.9 & 24.21/0.9 & 25.05/0.9 \\
\hline
1000 & Proposed & 7.04/1.6 & --- & --- & --- & --- & 7.15/1.7 & --- & --- & --- & --- \\
 & LM-CMA & 6.08/1.8 & --- & --- & --- & --- & 6.9/1.8 & --- & --- & --- & --- \\
\hline
  &      & \multicolumn{5}{c|}{2 cores, additional complexity 400} & \multicolumn{5}{c}{2 cores, additional complexity 600} \\\hline
100 & Proposed & 0.92/1.9 & 2.44/1.9 & --- & --- & --- & 1.9/1.9 & 4.81/1.9 & --- & --- & --- \\
 & LM-CMA & 0.59/1.9 & 2.9/1.9 & 78.27/1.9 & 80.25/1.9 & 82.35/1.9 & 1.51/1.9 & 6.76/1.9 & 179.06/1.9 & 183.73/1.9 & 188.64/1.9 \\
\hline
300 & Proposed & 2.81/1.9 & --- & --- & --- & --- & 5.84/1.9 & --- & --- & --- & --- \\
 & LM-CMA & 2.92/1.8 & 136.75/1.7 & --- & --- & --- & 5.82/1.9 & 459.66/1.8 & 252.88/1.7 & 180.88/0.9 & 187.57/0.9 \\
\hline
1000 & Proposed & 12.46/1.8 & --- & --- & --- & --- & 19.8/1.9 & --- & --- & --- & --- \\
 & LM-CMA & 9.5/1.8 & --- & --- & --- & --- & 28.82/1.9 & --- & --- & --- & --- \\
\hline
  &      & \multicolumn{5}{c|}{4 cores, additional complexity 0} & \multicolumn{5}{c}{4 cores, additional complexity 200} \\\hline
100 & Proposed & 0.13/1.8 & 0.31/1.8 & --- & --- & --- & 0.14/3.4 & 0.34/3.4 & --- & --- & --- \\
 & LM-CMA & 0.04/2.6 & 1.23/2.5 & 5.12/2.3 & 5.24/2.3 & 5.38/2.3 & 0.1/3.4 & 0.38/3.5 & 12.53/3.5 & 12.85/3.5 & 13.16/3.5 \\
\hline
300 & Proposed & 0.75/2.1 & --- & --- & --- & --- & 0.66/3.0 & --- & --- & --- & --- \\
 & LM-CMA & 0.34/2.6 & 33.76/2.6 & 16.01/1.2 & 16.38/1.2 & --- & 0.51/3.4 & 36.05/3.2 & --- & --- & --- \\
\hline
1000 & Proposed & 6.51/2.1 & --- & --- & --- & --- & 5.59/2.5 & --- & --- & --- & --- \\
 & LM-CMA & 4.2/2.8 & --- & --- & --- & --- & 4.63/3.1 & --- & --- & --- & --- \\
\hline
  &      & \multicolumn{5}{c|}{4 cores, additional complexity 400} & \multicolumn{5}{c}{4 cores, additional complexity 600} \\\hline
100 & Proposed & 0.37/3.9 & 0.88/3.9 & --- & --- & --- & 0.82/3.9 & 2.15/3.9 & --- & --- & --- \\
 & LM-CMA & 0.32/3.6 & 2.93/3.6 & 40.15/3.6 & 41.13/3.6 & 42.09/3.6 & 0.77/3.6 & 3.03/3.7 & 90.25/3.6 & 92.21/3.6 & 94.42/3.6 \\
\hline
300 & Proposed & 1.21/3.7 & --- & --- & --- & --- & 2.29/3.9 & --- & --- & --- & --- \\
 & LM-CMA & 2.03/3.5 & 86.35/2.9 & --- & --- & --- & 3.47/3.6 & 310.07/3.6 & --- & --- & --- \\
\hline
1000 & Proposed & 6.83/3.0 & --- & --- & --- & --- & 9.83/3.5 & --- & --- & --- & --- \\
 & LM-CMA & 10.33/3.4 & --- & --- & --- & --- & 18.41/3.5 & --- & --- & --- & --- \\
\hline
  &      & \multicolumn{5}{c|}{6 cores, additional complexity 0} & \multicolumn{5}{c}{6 cores, additional complexity 200} \\\hline
100 & Proposed & 0.16/1.5 & 0.4/1.4 & --- & --- & --- & 0.16/4.0 & 0.37/4.2 & --- & --- & --- \\
 & LM-CMA & 0.04/3.1 & 0.44/3.0 & 5.33/2.4 & 5.44/2.4 & 5.57/2.3 & 0.08/4.8 & 0.79/4.9 & 11.45/4.9 & 11.77/4.9 & 12.06/4.9 \\
\hline
300 & Proposed & 0.9/1.7 & --- & --- & --- & --- & 0.75/3.5 & --- & --- & --- & --- \\
 & LM-CMA & 0.34/2.6 & 34.74/2.6 & 39.48/2.5 & 40.67/2.5 & 19.86/1.2 & 0.33/4.5 & 26.4/4.6 & 29.54/4.4 & 30.36/4.4 & 30.95/4.4 \\
\hline
1000 & Proposed & 6.98/2.0 & --- & --- & --- & --- & 6.5/2.4 & --- & --- & --- & --- \\
 & LM-CMA & 3.78/3.2 & --- & --- & --- & --- & 4.17/3.8 & --- & --- & --- & --- \\
\hline
  &      & \multicolumn{5}{c|}{6 cores, additional complexity 400} & \multicolumn{5}{c}{6 cores, additional complexity 600} \\\hline
100 & Proposed & 0.29/5.7 & 0.73/5.8 & --- & --- & --- & 0.62/5.9 & 1.68/5.9 & --- & --- & --- \\
 & LM-CMA & 0.3/5.1 & 2.44/5.1 & 38.03/5.1 & 39.0/5.1 & 39.91/5.1 & 0.76/5.3 & 6.53/5.3 & 86.33/5.2 & 88.47/5.2 & 90.83/5.2 \\
\hline
300 & Proposed & 0.96/5.1 & --- & --- & --- & --- & 1.93/5.7 & --- & --- & --- & --- \\
 & LM-CMA & 0.96/5.1 & 79.83/5.1 & 37.55/4.8 & 39.06/4.8 & 40.67/4.8 & 2.1/5.3 & 150.7/5.1 & 35.79/4.4 & 36.47/4.4 & --- \\
\hline
1000 & Proposed & 6.21/4.0 & --- & --- & --- & --- & 7.87/4.9 & --- & --- & --- & --- \\
 & LM-CMA & 7.73/4.2 & --- & --- & --- & --- & 14.87/4.7 & --- & --- & --- & --- \\
\hline
  &      & \multicolumn{5}{c|}{8 cores, additional complexity 0} & \multicolumn{5}{c}{8 cores, additional complexity 200} \\\hline
100 & Proposed & 0.17/1.3 & 0.42/1.2 & --- & --- & --- & 0.2/3.6 & 0.47/3.6 & --- & --- & --- \\
 & LM-CMA & 0.05/2.3 & 1.24/2.2 & 5.58/2.1 & 5.69/2.1 & 5.83/2.1 & 0.08/5.8 & 1.04/5.8 & 9.28/5.9 & 9.47/5.9 & 9.7/5.9 \\
\hline
300 & Proposed & 0.94/1.7 & --- & --- & --- & --- & 0.87/2.8 & --- & --- & --- & --- \\
 & LM-CMA & 0.34/2.5 & 44.85/2.4 & 41.63/2.4 & 40.6/2.4 & 34.76/2.4 & 0.34/5.5 & 29.96/5.6 & 26.72/5.2 & 24.77/5.2 & 12.68/2.6 \\
\hline
1000 & Proposed & 7.22/2.0 & --- & --- & --- & --- & 6.74/2.3 & --- & --- & --- & --- \\
 & LM-CMA & 3.92/3.2 & --- & --- & --- & --- & 4.54/4.1 & --- & --- & --- & --- \\
\hline
  &      & \multicolumn{5}{c|}{8 cores, additional complexity 400} & \multicolumn{5}{c}{8 cores, additional complexity 600} \\\hline
100 & Proposed & 0.22/7.1 & 0.55/7.2 & --- & --- & --- & 0.49/7.7 & 1.17/7.8 & --- & --- & --- \\
 & LM-CMA & 0.24/6.6 & 1.23/6.7 & 32.18/6.6 & 32.95/6.6 & 33.67/6.6 & 0.71/6.9 & 9.63/6.9 & 87.36/6.8 & 88.94/6.8 & 91.39/6.8 \\
\hline
300 & Proposed & 0.88/6.1 & --- & --- & --- & --- & 1.3/7.3 & --- & --- & --- & --- \\
 & LM-CMA & 0.89/6.4 & 66.77/6.2 & 36.36/5.5 & 37.58/5.4 & 38.56/5.4 & 2.43/6.8 & 233.21/6.5 & --- & --- & --- \\
\hline
1000 & Proposed & 6.35/4.7 & --- & --- & --- & --- & 6.8/5.9 & --- & --- & --- & --- \\
 & LM-CMA & 7.4/5.4 & --- & --- & --- & --- & 15.54/5.8 & --- & --- & --- & --- \\
\hline
\end{tabular}
\end{center}
\end{table*}
\begin{table*}[t]
\begin{center}
\caption{
Experiment results for the Ellipsoid function.
Legend: ``Proposed'' is the asynchronous modification of LM-CMA-ES described in this paper,
``LM-CMA'' is the original LM-CMA-ES implementation. The table entries consist of
the median wall-clock time to reach the corresponding threshold (in seconds)
and of the median processor load value, separated by a forward slash.
}\label{main-ellipsoid}
\setlength{\tabcolsep}{0.6em}
\begin{tabular}{cc|ccccc|ccccc}\hline
n & Algo & \multicolumn{5}{c|}{Thresholds} & \multicolumn{5}{c}{Thresholds} \\\cline{3-12}
 & & $3000$ & $100$ & $10^{-2}$ & $3 \cdot 10^{-6}$ & $10^{-10}$ & $3000$ & $100$ & $10^{-2}$ & $3 \cdot 10^{-6}$ & $10^{-10}$ \\\hline
  &      & \multicolumn{5}{c|}{2 cores, additional complexity 0} & \multicolumn{5}{c}{2 cores, additional complexity 200} \\\hline
100 & Proposed & 0.02/1.6 & 0.06/1.6 & --- & --- & --- & 0.04/1.8 & 0.1/1.8 & --- & --- & --- \\
 & LM-CMA & 0.02/1.7 & 0.37/1.7 & 1.89/1.7 & 3.4/1.7 & 4.92/1.7 & 0.05/1.8 & 0.93/1.9 & 4.57/1.9 & 8.54/1.8 & 12.49/1.8 \\
\hline
300 & Proposed & 0.32/1.7 & 0.58/1.7 & --- & --- & --- & 0.36/1.8 & 0.52/1.7 & --- & --- & --- \\
 & LM-CMA & 1.43/1.7 & 13.11/1.8 & 48.54/1.8 & 84.02/1.8 & 119.75/1.8 & 1.54/1.8 & 16.88/1.8 & 63.1/1.8 & 110.26/1.8 & 157.16/1.8 \\
\hline
1000 & Proposed & 7.3/1.8 & 5.66/0.9 & --- & --- & --- & 6.95/1.8 & 5.12/0.9 & --- & --- & --- \\
 & LM-CMA & 181.06/1.9 & 660.1/1.9 & --- & --- & --- & 185.05/1.9 & 660.87/1.9 & --- & --- & --- \\
\hline
  &      & \multicolumn{5}{c|}{2 cores, additional complexity 400} & \multicolumn{5}{c}{2 cores, additional complexity 600} \\\hline
100 & Proposed & 0.14/1.9 & 0.39/1.9 & --- & --- & --- & 0.29/1.9 & 0.9/1.9 & --- & --- & --- \\
 & LM-CMA & 0.19/1.9 & 3.23/1.9 & 17.41/1.9 & 32.68/1.9 & 47.76/1.9 & 0.6/1.9 & 7.49/1.9 & 40.08/1.9 & 73.33/1.9 & 108.08/1.9 \\
\hline
300 & Proposed & 0.67/1.9 & 0.72/1.9 & --- & --- & --- & 1.44/1.9 & 2.0/1.9 & --- & --- & --- \\
 & LM-CMA & 4.07/1.9 & 41.38/1.9 & 158.47/1.9 & 276.14/1.9 & 387.09/1.9 & 8.48/1.9 & 90.54/1.9 & 344.16/1.9 & 603.95/1.9 & 844.17/1.8 \\
\hline
1000 & Proposed & 7.78/1.9 & --- & --- & --- & --- & 9.72/1.9 & --- & --- & --- & --- \\
 & LM-CMA & 244.14/1.9 & 893.87/1.9 & --- & --- & --- & 319.63/1.9 & 201.08/1.9 & --- & --- & --- \\
\hline
  &      & \multicolumn{5}{c|}{4 cores, additional complexity 0} & \multicolumn{5}{c}{4 cores, additional complexity 200} \\\hline
100 & Proposed & 0.03/1.7 & 0.08/1.4 & --- & --- & --- & 0.04/3.2 & 0.07/2.9 & --- & --- & --- \\
 & LM-CMA & 0.02/2.8 & 0.28/2.7 & 1.43/2.6 & 2.61/2.5 & 3.74/2.5 & 0.04/3.5 & 0.64/3.6 & 3.36/3.5 & 6.1/3.5 & 8.82/3.5 \\
\hline
300 & Proposed & 0.26/2.6 & 0.46/2.5 & --- & --- & --- & 0.24/3.0 & 0.4/2.9 & --- & --- & --- \\
 & LM-CMA & 0.9/2.9 & 8.29/2.9 & 30.65/2.9 & 53.38/2.9 & 76.15/2.9 & 0.86/3.5 & 8.99/3.5 & 34.29/3.5 & 59.95/3.5 & 86.08/3.5 \\
\hline
1000 & Proposed & 4.35/3.3 & 6.81/3.3 & --- & --- & --- & 4.24/3.2 & 6.27/3.1 & --- & --- & --- \\
 & LM-CMA & 97.71/3.7 & 357.67/3.7 & --- & --- & --- & 104.96/3.4 & 395.21/3.3 & --- & --- & --- \\
\hline
  &      & \multicolumn{5}{c|}{4 cores, additional complexity 400} & \multicolumn{5}{c}{4 cores, additional complexity 600} \\\hline
100 & Proposed & 0.07/3.8 & 0.18/3.8 & --- & --- & --- & 0.2/3.9 & 0.37/3.9 & --- & --- & --- \\
 & LM-CMA & 0.13/3.6 & 2.02/3.6 & 10.49/3.6 & 19.14/3.6 & 27.64/3.6 & 0.36/3.7 & 4.45/3.6 & 24.19/3.6 & 43.76/3.6 & 62.9/3.6 \\
\hline
300 & Proposed & 0.41/3.7 & 0.63/3.5 & --- & --- & --- & 0.73/3.9 & --- & --- & --- & --- \\
 & LM-CMA & 2.23/3.6 & 24.57/3.6 & 93.33/3.6 & 161.2/3.6 & 228.31/3.6 & 6.11/3.4 & 58.09/3.4 & 234.18/3.3 & 400.76/3.4 & 570.06/3.3 \\
\hline
1000 & Proposed & 5.0/3.4 & --- & --- & --- & --- & 5.54/3.5 & --- & --- & --- & --- \\
 & LM-CMA & 134.38/3.4 & 495.07/3.4 & --- & --- & --- & 187.61/3.6 & 681.65/3.6 & --- & --- & --- \\
\hline
  &      & \multicolumn{5}{c|}{6 cores, additional complexity 0} & \multicolumn{5}{c}{6 cores, additional complexity 200} \\\hline
100 & Proposed & 0.04/1.3 & 0.1/1.1 & --- & --- & --- & 0.04/3.5 & 0.08/2.4 & --- & --- & --- \\
 & LM-CMA & 0.02/3.5 & 0.3/3.1 & 1.45/2.7 & 2.61/2.8 & 3.77/2.8 & 0.03/4.7 & 0.46/4.9 & 2.74/4.9 & 5.0/4.9 & 7.26/4.9 \\
\hline
300 & Proposed & 0.34/2.3 & 0.6/2.1 & --- & --- & --- & 0.3/3.3 & 0.2/1.1 & --- & --- & --- \\
 & LM-CMA & 0.88/3.2 & 7.98/3.2 & 29.82/3.2 & 51.71/3.2 & 73.57/3.2 & 0.93/4.5 & 8.04/4.5 & 31.01/4.4 & 53.49/4.4 & 76.14/4.4 \\
\hline
1000 & Proposed & 3.9/4.0 & --- & --- & --- & --- & 3.64/4.1 & --- & --- & --- & --- \\
 & LM-CMA & 89.57/4.5 & 322.84/4.5 & --- & --- & --- & 79.45/4.9 & 289.77/4.8 & --- & --- & --- \\
\hline
  &      & \multicolumn{5}{c|}{6 cores, additional complexity 400} & \multicolumn{5}{c}{6 cores, additional complexity 600} \\\hline
100 & Proposed & 0.08/5.3 & 0.15/5.1 & --- & --- & --- & 0.08/5.3 & 0.15/5.1 & --- & --- & --- \\
 & LM-CMA & 0.11/5.1 & 1.42/5.2 & 7.73/5.1 & 14.54/5.1 & 21.31/5.1 & 0.24/5.3 & 3.36/5.3 & 17.92/5.2 & 33.2/5.2 & 48.78/5.2 \\
\hline
300 & Proposed & 0.36/4.7 & 0.57/4.4 & --- & --- & --- & 0.36/4.7 & 0.57/4.4 & --- & --- & --- \\
 & LM-CMA & 2.26/5.1 & 20.84/5.0 & 80.58/5.1 & 138.97/5.0 & 198.07/5.0 & 4.16/5.3 & 39.42/5.3 & 146.97/5.3 & 251.66/5.3 & 361.75/5.3 \\
\hline
1000 & Proposed & 4.16/4.5 & 2.74/2.0 & --- & --- & --- & 4.16/4.5 & 2.74/2.0 & --- & --- & --- \\
 & LM-CMA & 97.49/5.0 & 359.01/5.0 & --- & --- & --- & 161.67/4.6 & 613.68/4.6 & --- & --- & --- \\
\hline
  &      & \multicolumn{5}{c|}{8 cores, additional complexity 0} & \multicolumn{5}{c}{8 cores, additional complexity 200} \\\hline
100 & Proposed & 0.04/1.3 & 0.1/1.0 & --- & --- & --- & 0.05/3.2 & 0.11/2.6 & --- & --- & --- \\
 & LM-CMA & 0.02/2.8 & 0.31/2.7 & 1.56/2.6 & 2.77/2.5 & 3.98/2.5 & 0.03/6.0 & 0.41/6.0 & 2.13/6.0 & 3.89/6.0 & 5.64/6.0 \\
\hline
300 & Proposed & 0.37/1.9 & 0.64/1.8 & --- & --- & --- & 0.33/2.7 & 0.54/2.3 & --- & --- & --- \\
 & LM-CMA & 0.97/3.1 & 9.13/3.0 & 35.07/3.0 & 60.18/3.0 & 86.06/3.0 & 0.9/5.3 & 7.91/5.0 & 29.44/5.0 & 51.42/4.9 & 72.73/4.9 \\
\hline
1000 & Proposed & 3.95/3.9 & --- & --- & --- & --- & 3.95/4.2 & --- & --- & --- & --- \\
 & LM-CMA & 85.91/5.1 & 319.85/5.1 & --- & --- & --- & 72.46/5.6 & 269.39/5.6 & --- & --- & --- \\
\hline
  &      & \multicolumn{5}{c|}{8 cores, additional complexity 400} & \multicolumn{5}{c}{8 cores, additional complexity 600} \\\hline
100 & Proposed & 0.07/6.6 & 0.15/5.2 & --- & --- & --- & 0.14/7.3 & 0.22/6.9 & --- & --- & --- \\
 & LM-CMA & 0.1/6.6 & 1.14/6.7 & 6.84/6.6 & 12.26/6.5 & 17.53/6.5 & 0.22/7.0 & 3.75/6.9 & 18.11/6.8 & 33.17/6.8 & 48.28/6.8 \\
\hline
300 & Proposed & 0.36/5.0 & 0.57/3.4 & --- & --- & --- & 0.54/7.1 & 0.6/5.4 & --- & --- & --- \\
 & LM-CMA & 1.65/6.1 & 15.71/6.2 & 60.05/6.2 & 105.71/6.2 & 150.66/6.2 & 3.26/6.6 & 28.07/6.7 & 109.14/6.7 & 193.33/6.7 & 277.51/6.7 \\
\hline
1000 & Proposed & 3.72/5.0 & --- & --- & --- & --- & 4.28/6.1 & --- & --- & --- & --- \\
 & LM-CMA & 82.97/6.0 & 316.75/5.9 & --- & --- & --- & 141.83/5.4 & 546.15/5.3 & --- & --- & --- \\
\hline
\end{tabular}
\end{center}
\end{table*}

}

For the Sphere function (Table~\ref{main-sphere}), one can see that both algorithms reach the 
optimum of this problem.
The worst behavior of the proposed algorithm can be seen when the number of cores 
is high (e.g. 8) and the fitness function complexity is low. This corresponds to the case where 
the most time is spent on waiting for a critical section. The best behavior is demonstrated when 
the number of cores is high and the fitness function complexity is also high. Under these 
conditions, the proposed algorithm, while using an update rule which is generally worse,
reaches the optimum faster due to more efficient usage of multiple cores.

For the Rastrigin function (Table~\ref{main-rastrigin}), the proposed algorithm always finds an 
optimum, but the original 
LM-CMA-ES never does that, which was surprising. This change can be attributed to the update rule 
as well, but in this case a simpler rule brings better results.

The Rosenbrock (Table~\ref{main-rosenbrock}) and the Ellipsoid (Table~\ref{main-ellipsoid}) 
functions are particularly hard for both algorithms. The proposed algorithm typically produces 
worse results, but in the case of high computational complexity of the fitness function it reaches 
the thresholds faster. This can be attributed both to the update rule (which seems to be too 
greedy for these problems) and to the efficient usage of multiple cores.

\section{Conclusion}\label{conclusion}

We presented our first attempt to implement an asynchronous version of the LM-CMA-ES algorithm. 
Our algorithm is the best when the number of cores is high and the computation complexity of the 
fitness function is also high. It uses a different update rule than the LM-CMA-ES algorithm
which generally is not very good, however, the efficient usage of multiple cores compensates for 
this fact on the Sphere benchmark function, and brings to some fitness functions thresholds faster
on the Rosenbrock and the Ellipsoid function. A surprising fact is that it performs on the 
Rastrigin problem much better than the original algorithm.

\iffullpaper{\newpage}
This work was partially financially supported by the Government of Russian Federation, Grant 074-U01.

\bibliographystyle{IEEEtran}
\bibliography{../../../bibliography}

\begin{thebibliography}{1}
\providecommand{\url}[1]{#1}
\csname url@samestyle\endcsname
\providecommand{\newblock}{\relax}
\providecommand{\bibinfo}[2]{#2}
\providecommand{\BIBentrySTDinterwordspacing}{\spaceskip=0pt\relax}
\providecommand{\BIBentryALTinterwordstretchfactor}{4}
\providecommand{\BIBentryALTinterwordspacing}{\spaceskip=\fontdimen2\font plus
\BIBentryALTinterwordstretchfactor\fontdimen3\font minus
  \fontdimen4\font\relax}
\providecommand{\BIBforeignlanguage}[2]{{%
\expandafter\ifx\csname l@#1\endcsname\relax
\typeout{** WARNING: IEEEtran.bst: No hyphenation pattern has been}%
\typeout{** loaded for the language `#1'. Using the pattern for}%
\typeout{** the default language instead.}%
\else
\language=\csname l@#1\endcsname
\fi
#2}}
\providecommand{\BIBdecl}{\relax}
\BIBdecl

\bibitem{bfgs}
D.~F. Shanno, ``Conditioning of quasi-{N}ewton methods for function
  minimization,'' \emph{Mathematics of Computation}, vol.~24, no. 111, pp.
  647--656, 1970.

\bibitem{hansen-2003}
N.~Hansen, S.~D. M{\"u}ller, and P.~Koumoutsakos, ``{R}educing the {T}ime
  {C}omplexity of the {D}erandomized {E}volution {S}trategy with {C}ovariance
  {M}atrix {A}daptation ({CMA-ES}),'' \emph{Evolutionary Computation}, vol.~11,
  no.~1, pp. 1--18, 2003.

\bibitem{numeric-cma-update}
O.~Krause and C.~Igel, ``A more efficient rank-one covariance matrix update for
  evolution strategies,'' in \emph{Proceedings of Foundations of Genetic
  Algorithms XIII}, 2015, pp. 129--136.

\bibitem{cma-es-diagonal}
R.~Ros and N.~Hansen, ``{A} {S}imple {M}odification in {CMA-ES} {A}chieving
  {L}inear {T}ime and {S}pace {C}omplexity,'' in \emph{Parallel Problem Solving
  from Nature X}, ser. Lecture Notes in Computer Science.\hskip 1em plus 0.5em
  minus 0.4em\relax Springer, 2008, no. 5199, pp. 296--305.

\bibitem{glasmachers-async-nes}
T.~Glasmachers, ``A {N}atural {E}volution {S}trategy with {A}synchronous
  {S}trategy {U}pdates,'' in \emph{Proceeding of Genetic and Evolutionary
  Computation Conference}.\hskip 1em plus 0.5em minus 0.4em\relax ACM, 2013,
  pp. 431--437.

\bibitem{natural-es}
D.~Wierstra, T.~Schaul, T.~Glasmachers, Y.~Sun, J.~Peters, and J.~Schmidhuber,
  ``Natural evolution strategies,'' \emph{Journal of Machine Learning
  Research}, vol.~15, pp. 949--980, 2014.

\bibitem{lm-cma-es}
I.~Loshchilov, ``A {C}omputationally {E}fficient {L}imited {M}emory {CMA-ES}
  for {L}arge {S}cale {O}ptimization,'' in \emph{Proceeding of Genetic and
  Evolutionary Computation Conference}.\hskip 1em plus 0.5em minus 0.4em\relax
  ACM, 2014, pp. 397--404.

\bibitem{mendel-2015-async-cma}
V.~Arkhipov and M.~Buzdalov, ``An asynchronous implementation of the limited
  memory {CMA-ES}: First results,'' in \emph{Proceedings of International
  Conference on Soft Computing MENDEL}, 2015, pp. 37--40.

\end{thebibliography}

\end{document}